\documentclass{article}

\usepackage{bm}
\usepackage{microtype}
\usepackage{graphicx}
\usepackage{booktabs} %
\usepackage{adjustbox} %

\usepackage[hidelinks,colorlinks,hyperfootnotes=false]{hyperref}
\usepackage{xurl}
\usepackage[normalem]{ulem}

\usepackage[accepted]{icml2025}

\usepackage{amsmath}
\usepackage{amssymb}
\usepackage{mathtools}
\usepackage{makecell}
\usepackage{amsthm}
\usepackage{tcolorbox} %
\usepackage{multirow}
\usepackage{pifont}
\usepackage{tablefootnote}
\usepackage{subcaption}
\usepackage{tikz}
\usepackage{graphicx} %

\usepackage[capitalize,noabbrev]{cleveref}

\theoremstyle{plain}

\theoremstyle{definition}

\theoremstyle{remark}

\usepackage[normalem]{ulem}

\usepackage[colorinlistoftodos,textsize=scriptsize]{todonotes}

\newcommand\blfootnote[1]{%
  \begingroup
  \renewcommand\thefootnote{}\footnote{#1}%
  \addtocounter{footnote}{-1}%
  \endgroup
}

\newcommand{\datasetname}{\textbf{\texttt{STAMP}}}
\newcommand{\datasetnamespace}{\textbf{\texttt{STAMP}} }

\newcommand{\datasetdesc}{\textbf{\datasetname}~(\textbf{S}potting \textbf{T}raining \textbf{A}rtifacts through water\textbf{M}arked \textbf{P}airs)}

\icmltitlerunning{\texttt{STAMP} Your Content: Proving Dataset Membership via Watermarked Rephrasings}

\begin{document}

\twocolumn[

\newcommand{\stamplogo}{\raisebox{-0.6em}{\includegraphics[height=1.7em]{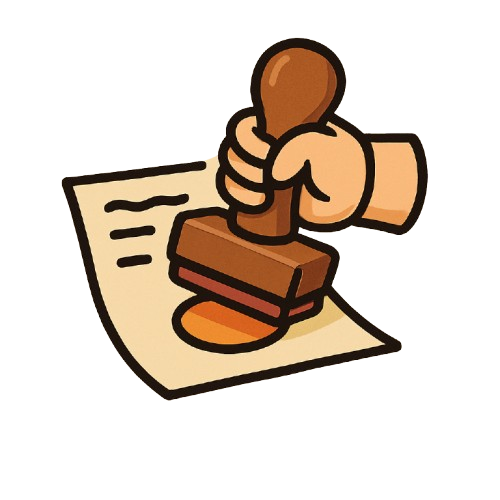}}~\hspace{-3.5pt}~\texttt{STAMP}}

\icmltitle{\stamplogo{} Your Content:\\Proving Dataset Membership via Watermarked Rephrasings}

\icmlsetsymbol{equal}{*}

\begin{icmlauthorlist}
\icmlauthor{Saksham Rastogi}{yyy}
\icmlauthor{Pratyush Maini}{comp,zzz}
\icmlauthor{Danish Pruthi}{yyy}
\end{icmlauthorlist}

\icmlaffiliation{yyy}{Indian Institute of Science}
\icmlaffiliation{comp}{Carnegie Mellon University}
\icmlaffiliation{zzz}{DatologyAI}

\icmlcorrespondingauthor{Saksham Rastogi}{iitdsaksham@gmail.com}
\icmlcorrespondingauthor{Pratyush Maini}{pratyushmaini.@cmu.edu}

\icmlkeywords{Machine Learning, LLMs, Watermark, Membership Inference, Dataset Inference, Natural Language Processing, Generative AI}

\vskip 0.3in
]

\printAffiliationsAndNotice{}

\begin{abstract}

Given how large parts of publicly available text 
are crawled to pretrain large language models~(LLMs), 
data creators increasingly worry about the 
inclusion of their proprietary 
data for model training 
without attribution or licensing.
Their concerns are also shared by benchmark curators 
whose test-sets might be compromised.
In this paper, we present 
\datasetname,
a framework for detecting dataset membership---i.e., determining
the inclusion of a dataset in the pretraining corpora of 
LLMs.
Given an original piece of content, 
our proposal involves first generating multiple rephrases, 
each embedding a watermark with a unique secret key.
One version is to be released publicly, while others are to be kept private. 
Subsequently, creators can compare model likelihoods 
between public and private versions 
using paired statistical tests to prove membership.
We show that our framework can successfully detect contamination 
across four benchmarks which appear only once 
in the training data and constitute less than
$0.001$\% of the total tokens, 
outperforming several contamination detection 
and dataset inference baselines.
We verify that 
\datasetnamespace
preserves 
both the semantic meaning and
utility of 
the original data. 
We apply \datasetnamespace to two real-world scenarios
to confirm the inclusion of paper abstracts and 
blog articles in the pretraining corpora.

\end{abstract}

\section{Introduction} 
\begin{figure*}[t]
    \centering
    \includegraphics[width=0.98\textwidth]{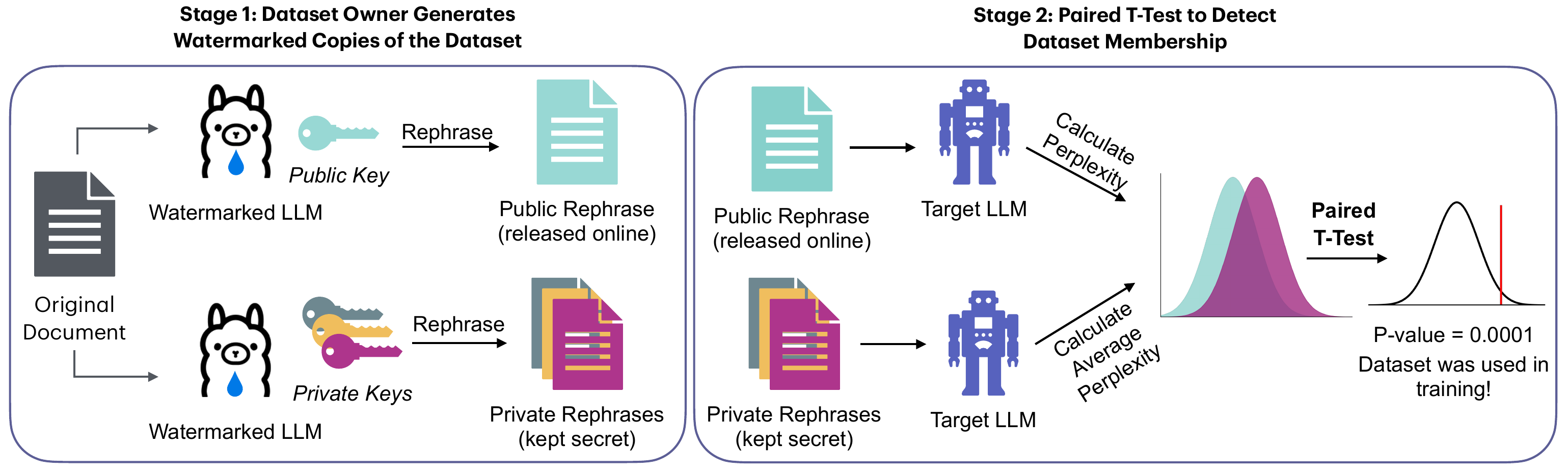}
    \caption{\textbf{Overview of} \datasetname. 
    \textbf{Stage 1:} \emph{Create Watermarked Copies of the Dataset.}
    We use a watermarked LLM to generate multiple 
    rephrased versions of their original dataset, 
    each uniquely watermarked using a distinct key.
    The version watermarked with the public key 
    is released publicly on the internet, 
    while other watermarked versions are kept private.
    \textbf{Stage 2:} \emph{Prove Membership using a Paired T-Test.}
    To detect membership, we compute model perplexities
    over documents from both the public and private versions.
    Using these perplexity scores, we perform a paired t-test to 
    detect membership. 
    }
    
    \label{fig:llm-test-set-contamination-overview}
\end{figure*}

To train large language models, 
much of the available text
from the internet is crawled, 
allegedly including copyrighted material 
such as news articles and blogs \cite{Grynbaum2023LawsuiteOpenAINewYorkTimes, SilvermanLawsuitOpenAI}. 
Additionally,
some evaluation datasets, originally intended for 
benchmarking model performance, 
may be compromised---an issue 
prominently discussed as 
\emph{test-set contamination}
\citep{magar-schwartz-2022-data, jacovi2023stopuploadingtestdata, sainz-etal-2023-nlp}.
A recent study
reveals
concerning 
evidence 
that pretraining corpora 
contain several
key benchmarks \cite{elazar2024whatsbigdata}, 
and another demonstrates 
that impact of test set contamination 
has been underestimated in many 
prominent LLM releases~\cite{singh2024evaluation}.

On one hand, 
training language models 
on copyrighted material 
might violate legal standards, 
and on the other,
consuming test sets of machine learning 
benchmarks
might offer a false sense of progress.
Given the 
lack of regulations or incentives 
for model developers
to disclose contents of their pretraining corpora 
\cite{openai2024gpt4technicalreport, llama3modelcard, 2024anthropic},
it is critical 
to equip 
content creators 
with reliable 
tools to determine 
whether 
their content 
was included as a part of model training.
Especially,
third party 
approaches that can democratize
detecting dataset membership and
enable independent accountability.
\blfootnote{We make all our code, data and models available at \href{https://github.com/codeboy5/stamp}{github.com/codeboy5/STAMP}}

Some approaches for detecting dataset membership
embed random sequences in text or substitute 
characters with visually-similar unicodes \cite{wei2024proving}.
However, such alterations 
impair machine readability, indexing and 
retrieval---making them impractical for content creators.
More critically for benchmarks, such substitutions 
can alter tokenization, potentially compromising their utility for evaluation.
Other proposals 
rely on access to a
\emph{validation} set that is unseen by the target model 
and drawn from the same distribution as the original dataset---a requirement hard to meet in practice \citep{maini2024llmdatasetinferencedid}. 
Recently, 
\citet{oren2023provingtestsetcontamination}
suggest
comparing 
canonical ordering of test sets to random permutations,
but this strategy 
assumes 
large portions of datasets 
are processed together within a single context window 
during pretraining.
Most closely related to our proposal,
\citet{zhang2024pacostpairedconfidencesignificance}
use a statistical test to compare model confidence 
on original test instances and their rephrasings, 
assuming that the two distributions are identical---an assumption
we show does not hold (Table \ref{table::iid_check}).

In our work, we propose
\datasetdesc,
a practical approach 
allowing creators to 
detect dataset membership through a statistical test 
with a probabilistic interpretation~(Figure~\ref{fig:llm-test-set-contamination-overview}).
Our approach begins by taking the original content 
and generating multiple rephrased versions.
Each rephrased version is watermarked using a distinct
key for the hash function used in watermarking.
Content creators can then release 
one of the generations publicly, 
while keeping the others private.
A statistical test then evaluates 
the model likelihood of 
generating the public version 
against the private copies.
For models that were trained 
on the publicly available generations, 
we expect to observe 
higher model likelihoods for 
these generations
compared to their private counterparts.

Our work repurposes LLM watermarking to watermark documents
that considerably enhance the detection sensitivity of our statistical test. (This is different 
from watermarking \emph{models} themselves 
to prevent against model extraction attacks.)
Specifically, we leverage the KGW  watermarking scheme \cite{kirchenbauer2024reliabilitywatermarkslargelanguage},
which embeds detectable signals by steering generations towards a randomly chosen ``green'' subset of the vocabulary.

We empirically validate the effectiveness of our 
approach by continually 
pretraining the  
Pythia~1B model \citep{biderman2023pythiasuiteanalyzinglarge}
on deliberately contaminated pretraining data. 
We contaminate the pretraining corpus 
by injecting test examples from four 
different benchmarks.
Even with minimal contamination---that is, each test example appearing only once 
and each benchmark comprising less than $0.001$\% of the total training data---our approach
significantly outperforms existing methods, 
achieving statistically significant 
p-values across all contaminated benchmarks.
We also conduct a false positive analysis, 
wherein we apply 
our detection methodology to off-the-shelf pretrained LLMs that have not been exposed to
the watermarked benchmarks
and find that 
our test successfully  
denies their membership.
Moreover, our analysis reveals that watermarking substantially 
enhances detection sensitivity, improving statistical significance 
by up to three orders of magnitude.

To demonstrate \datasetname's effectiveness in 
detecting inclusion of copyrighted data in pretraining corpora,
we present two expository case studies 
where we apply \datasetnamespace to 
detect membership of paper abstracts and blog articles.
Our test achieves statistically significant p-values 
across these real-world scenarios.
To further ensure that our framework preserves content quality,
we conduct both automatic evaluations 
using GPT4~\citep{openai2024gpt4technicalreport} and a human study,
and find 
that \datasetnamespace maintains content quality.
These results 
highlight  
its utility
in protecting copyrighted material (for creators), and detecting contamination (for auditors).

\section{Preliminaries} \label{sec:problem_setup}

In this section, we begin by formalizing the problem 
of detecting membership of a dataset (\S\ref{subsec:formal_problem}) and
provide necessary background on 
watermarks for LLMs (\S\ref{subsec::llm_watermarks}).

\subsection{Dataset Membership} \label{subsec:formal_problem}

The problem of dataset membership~\citep{maini2021dataset} 
aims to determine  
whether a dataset $X$
has been included in the pretraining data $D_{\text{train}}$ 
of a language model $\theta$.
We operate under a \emph{gray-box} setting,
where we can compute token probabilities for any sequence 
$S$ but have no access to the pretraining data or model weights.
Formally detecting membership of a dataset
can be posed
as a hypothesis test with the goal to 
distinguish between the following hypothesis:
\begin{itemize}
\item $\bm{H_0}$: $\theta$ is independent of $X$ (no membership)
\item $\bm{H_1}$: $\theta$ is dependent on $X$ (membership),
\end{itemize}
where we treat $\theta$ as a random variable 
whose randomness arises from the sampling of 
the pretraining dataset $D_{\text{train}}$
(which may or may not include $X$).
Framing 
membership inference~\citep{shokri2017membershipinferenceattacksmachine}
as hypothesis testing provides 
statistical guarantees on the false detection rate.

\textbf{Our focus} is on building statistical tests that can reliably
detect dataset membership in language models.
We aim to develop methods 
that make minimal assumptions about the format or nature of data---be it machine learning benchmarks, newsletters, or books.

\subsection{Watermarks for LLMs} \label{subsec::llm_watermarks}

Watermarking techniques for LLMs embed subtle
but distinctive patterns within generated text
that are imperceptible to humans but algorithmically detectable.
For our framework, we utilize the prominent KGW scheme \cite{kirchenbauer2024reliabilitywatermarkslargelanguage}.
KGW scheme uses a hash function that takes the context
(preceding tokens) and a hash key $h$ to partition the vocabulary $V$ into
two disjoint sets at each generation step: a green list $G$ and a red list $R$.  

To embed a watermark, the scheme biases the model's 
next-token probabilities
by adding 
$\delta$ 
($\delta > 0$)
to the logits of tokens in the green list.
Specifically, if $l^{(t)}_k$ denotes the original logit for token $k$ at position $t$,
then the modified logits are given by:
\begin{equation}
\hat{l}_k^{~(t)} \leftarrow l^{(t)}_k + \delta 1\!\!1\big[k \in G\big].
\label{eq:watermark_prob}
\end{equation}

\section{
STAMP: \uline{S}potting \uline{T}raining 
\uline{A}rtifacts through Water\uline{m}arked \uline{P}airs} \label{sec:methodology}

We introduce
\datasetname,
a practical and 
principled framework that enables content creators 
to reliably detect whether their content was included in LLM 
pretraining data.
Our approach builds on a key insight: 
if an LLM consistently 
prefers documents watermarked with a specific key
(e.g., the key used for the publicly available version)
over semantically equivalent content with distinct watermarks,
then the model must have seen the preferred documents during 
pretraining.
In this section, we detail how \datasetnamespace leverages this insight 
to create a robust statistical framework for membership detection.
\datasetnamespace consists of two stages:
(1) a process for content creators
to release watermarked content (\S\ref{method:stage0}) and 
(2) a paired statistical test to detect downstream dataset membership (\S\ref{method:stage1}).

\subsection{Watermarking Datasets} \label{method:stage0}

The first stage of our approach involves 
generating multiple watermarked
versions of a 
dataset through rephrasing. 
For a given dataset $X$, we employ an 
open-weights
instruction-tuned 
LLM to generate rephrases.
For each document $q$ in the original dataset,
we create a public version (denoted as $q^{'}$),
where the rephrase is watermarked using a designated 
public key as the \emph{hash key}.
Additionally, we generate $m$ 
private versions (denoted as $q_1^{''}, q_2^{''}, \dots, q_m^{''}$),
where each generation is watermarked 
using a distinct private key
as the hash key.
The public version is 
released online,
while the private versions are
kept confidential.
Crucially, due to the design of our test
relying on pairwise comparisons at a document level (\S\ref{method:stage1}),
each document $q$ in a dataset $X$ 
can use a different set of hash keys.
This ensures that introducing watermarking 
during the rephrasing stage does not alter
the token distribution of the dataset $X$ and,
importantly, preserves the overall token distribution 
of the internet data.

\paragraph{LLM Watermarks as Sampled Markers.}
While watermarking is traditionally intended 
for attributing generated text to a specific LLM, 
our motivation diverges from this original purpose.
First, we leverage LLM watermarking as a mechanism to 
embed distinct signals into the rephrases
through the use of distinct hash keys.
The randomness in both our hash key 
selection and the watermarking
process itself enables us to 
frame the detection problem as
hypothesis testing. 
Under the null hypothesis $\bm{H_0}$ (no membership),
the target model shouldn't 
favor content watermarked with
any particular key.
Second, the watermarking process itself 
introduces subtle perturbations that 
increase sequence perplexity, 
which has been empirically shown 
to enhance memorization during 
training \cite{meeus2024copyright}, 
further amplifying our 
ability to detect membership.

\subsection{Detecting Dataset Membership} \label{method:stage1}

To detect membership, we leverage 
the insight that
under the null hypothesis $\bm{H_0}$ (no membership),
the model should not exhibit any
systematic preference towards any of the semantically equivalent paraphrases of documents
that are watermarked with distinct keys---the public version of the dataset
and privately held versions of the dataset.
This follows from the randomness inherent in our 
selection of keys and nature of watermark we employ.
We formalize this intuition through a statistical testing framework.

For each document $q$, we compute the perplexity difference $d_i$ between its 
public version $q_i^{'}$ and private version $q_i^{''}$ that form a pair $(q_i^{'}, q_i^{''})$:
\begin{equation}
\label{eq:paired_differences}
d_i  = \textit{PPL}_{\theta}(q_i^{'}) - \textit{PPL}_{\theta}(q_i^{''}).
\end{equation}
Prior to applying the paired t-test, we modify 
the top 5\% outliers by clipping their values. 
This prevents issues where the test can 
become ineffective due to a few outlier samples.
Under the alternative hypothesis $\bm{H_1}$,
we expect these differences 
to be 
negative
on average, 
indicating lower perplexity for public versions. 
We evaluate this using a one-sided paired t-test statistic:
\begin{equation}
\label{eq:t_test_statistic}
t = \frac{\bar{d}}{s_d / \sqrt{n}} ,
\end{equation}
where $\bar{d}$ and $s_d$ are the mean and standard deviation of the differences 
across the collection of documents respectively and $n$ 
is the number of documents~\citep{student1908probable}.
The one-sided p-value specifically tests for $\bar{d} < 0$, 
following our 
alternative hypothesis
that exposure during training leads 
to lower perplexity on public versions. 
Paired tests provide higher statistical power, 
enabling detecting membership 
even with a smaller collection of documents ($n$), 
as we show in our experiments.

\paragraph{Multiple Private Keys.}

In practice, we empirically observe that models may exhibit inherent biases
at an individual document level,
occasionally assigning lower perplexity to 
specific private rephrases ($q{''}$) independent of membership. 
To make our detection robust against such biases, 
we propose using multiple private rephrases,
each watermarked with a distinct key.
Instead of comparing against a single private version, 
we test whether public rephrases exhibit lower perplexity 
compared to the average perplexity 
across $m$ different private rephrases: 
\begin{equation}
\label{eq:t_test_statistic_multi_avg}
d_i = \textit{PPL}_{\theta}(q_i^{'}) - 
\frac{1}{m} \sum_{j=1}^{j=m} \textit{PPL}_{\theta}(q_{i,j}^{''}),
\end{equation}

where $m$ is a hyperparameter known as \textit{private key count}, 
and $q_{i,j}^{''}$ represents the $j^{th}$ private rephrase of $i^{th}$ document.
Through controlled experiments, we analyze 
the effect of this hyperparameter on 
statistical strength of our test (\S\ref{subsec:test_parameters}).

\section{Experiments \& Results} \label{sec:experiments}

\begin{table*}[th]
\caption{\textbf{P-values for detecting \emph{test-set contamination} for different methods.} 
For LLM DI \cite{maini2024llmdatasetinferencedid}, 
\textit{same} refers to using rephrases of the benchmark questions 
as \emph{validation} set, while \textit{different} uses an 
entirely different set of unseen questions from the same benchmark as the \emph{validation} set.
\textbf{Bold} indicates statistically significant results ($p < 0.05$). Across all the four benchmarks, our approach results in lower p-values compared to other approaches (lower is better).}
\begin{center}
    \begin{center}
    \begin{sc}
    \begin{tabular}{lcccc}
    \toprule
    & \multicolumn{4}{c}{Benchmark ($\downarrow$)} \\
    \cmidrule(lr){2-5}
    Method           & TriviaQA & Arc-C & MMLU   & GSM8K  \\
    \midrule
    PaCoST \cite{zhang2024pacostpairedconfidencesignificance}   &   \textbf{1.6e-3} &    0.33      &  0.19     &  0.21     \\
    \midrule
    LLM DI \cite{maini2024llmdatasetinferencedid}  (\textit{same})        &      0.43    &   0.31    &   0.46    &   0.30    \\
    LLM DI \cite{maini2024llmdatasetinferencedid}  (\textit{different})   &   \textbf{0.02}  &   0.53    &   \textbf{0.03}  &   0.71 \\
    \midrule
    \datasetnamespace (w/o paired tests)        &   {0.14}       &   {0.07}   &   {0.08}   &  \textbf{0.02}   \\
    \datasetnamespace (w/o watermarking)        &   \textbf{0.02}     &   \textbf{5.1e-3} &   \textbf{0.02} &  \textbf{1.4e-3}   \\
    \datasetnamespace                           &   \textbf{1.2e-4}     &  \textbf{2.8e-4}  &   \textbf{7.0e-4} & \textbf{6.6e-6}  \\
    \bottomrule
    \end{tabular}
    \end{sc}
    \end{center}
\end{center}
\label{table::main_results}
\end{table*}

To evaluate the ability of \datasetnamespace for membership detection,
we first focus on benchmark contamination---the inclusion of evaluation benchmarks
in the pretraining corpora of LLMs.
This setting presents unique challenges for membership detection.
First, benchmarks must maintain their utility as reliable indicators of 
progress, which constrains the modifications we can make prior to their release.
Second, benchmarks typically contain limited text 
compared to other content types (e.g., books or newsletters), 
making detection particularly challenging.

\subsection{Releasing Watermarked Test Sets} \label{sec:release_watermark_set}

We evaluate our approach using four widely-used benchmarks:
TriviaQA \cite{TriviaQA}, 
ARC-C \cite{ARC-C}, 
MMLU \cite{MMLU}, 
and GSM8K \cite{GSM8K}.
For each benchmark, we follow our proposed
methodology (\S\ref{method:stage0}) to 
generate
watermarked public and private paraphrases.
We use the instruction tuned Llama3-70B \cite{llama3modelcard} model
and a benchmark-agnostic prompt 
(provided in Appendix \ref{appendix:prompt_template})
to generate these rephrased copies.
For each benchmark, we randomly select 
one watermarked version to be the 
\emph{public} version.
Examples of the rephrased test instances 
are provided in Appendix \ref{appendix:rephrased_examples}.

\paragraph{Key Distinction.}
While rephrasing has been previously explored 
for detecting contamination \cite{zhang2024pacostpairedconfidencesignificance},
existing approaches typically compare human-written content against their LLM-generated rephrases,
overlooking a crucial confounding factor: language models exhibit 
systematic preferences for LLM-generated text over 
human-written content \cite{liu2023gevalnlgevaluationusing, Mishra2023CharacterizingLL, Laurito2024AIAB}.
This inherent bias undermines the reliability of statistical approaches that compare
human-written content with their LLM rephrasings, as any detected differences 
might stem from this general preference rather than training exposure.
To enable reliable statistical testing, it is crucial 
to control the data generating process for both versions being compared.
We address this by ensuring both our public and private versions 
are generated through the same process, differing only in their watermarking keys.
Given the random selection of keys, we expect no systematic preferences 
between versions unless one was seen during training.

We empirically validate that 
human-written content and its LLM-generated rephrasings 
are easily 
distinguishable 
(thus violating the expected IID requirement): 
a simple bag-of-words classifier obtains
$\text{AUROC} > 0.8$ on four out of five benchmarks,
whereas 
the classifier performs no better
than random chance when distinguishing between rephrasings watermarked with
different keys. Detailed analysis and classifier specifications are provided
in Appendix \ref{appendix:bagofwords}.

\subsection{Pretraining with Intentional Contamination} \label{subsec:exp_intentional_contam}

\paragraph{Setup.} \label{para:setup} 
To simulate downstream benchmark contamination as it occurs in real-world scenarios 
and evaluate the effectiveness of our test, 
we perform continual pretraining on the 1 billion parameter 
Pythia model \cite{biderman2023pythiasuiteanalyzinglarge} 
using an intentionally contaminated pretraining corpus.
The corpus is a combination of OpenWebText \cite{2023opencompass}
and \emph{public} watermarked version of the four benchmarks,
as mentioned in Section \ref{sec:release_watermark_set}.
Each test set accounts for less than \textbf{0.001\%} 
of the pretraining corpus, with exact sizes detailed in Table \ref{tab:contamination_sizes} in the appendix. 
All test sets in our experiments have a duplication rate of $1$ (denoting no duplication whatsoever), 
and the overall pretraining dataset comprises $6.7$ billion tokens.
Details of the exact training hyperparameters
are provided in Appendix \ref{appendix:pretraining_details}.

\paragraph{Baselines.}

We compare \datasetnamespace against two recent statistical approaches to detect membership:
PaCoST \cite{zhang2024pacostpairedconfidencesignificance} and
LLM DI \cite{maini2024llmdatasetinferencedid}. 
PaCoST employs a paired t-test that compares model confidence
on original and rephrased versions, while LLM DI aggregates
multiple membership inference attacks (MIAs) to perform statistical
testing. For LLM DI, which requires access to an unseen \emph{validation} set,
we evaluate two settings: (1) using private rephrases of the
publicly available dataset as the \emph{validation} set, and (2) using an
entirely different set of documents from the same distribution as the \emph{validation} set.

Additionally, we also evaluate state-of-the-art MIAs:
\textit{PPL} \cite{yeom2018privacy},
\textit{Zlib} \cite{carlini2021extracting},
\textit{Min-K} \cite{shi2024detectingpretrainingdatalarge},
\textit{Min-K++} \cite{zhang2024min} and
\textit{DC-PDD} \cite{zhang2024pretrainingdatadetectionlarge}.
Since MIAs rely on a non-trivial detection
threshold, we report AUROC scores across two settings:
(1) discriminating between public rephrases in training and
private rephrases of the same documents, and 
(2) discriminating between public rephrases in training 
and unseen documents from the same dataset.

\paragraph{Main Results.}

We compare \datasetnamespace and baseline methods in Table \ref{table::main_results}.
\datasetnamespace achieves statistically significantly low p-values (ranging from $10^{-4}$ to $10^{-6}$) across all
benchmarks, substantially outperforming existing methods.
In contrast, PaCoST detects contamination only on TriviaQA ($p \approx 10^{-3}$),
while LLM DI shows significance on just two benchmarks (TriviaQA and MMLU) 
even with access to validation data of extra test examples.

In our experiments, all MIA methods achieve an AUROC score of 
$\approx 0.5$ across all benchmarks, 
indicating performance no better 
than random guessing.
Detailed MIA results and analysis are presented 
in Table \ref{table::mia_metrics}.

\begin{table}[t]
\caption{ \textbf{False positive analysis.}
\emph{Pythia Uncontaminated} denotes the p-values
on a pretrained Pythia model that has not been contaminated.
\emph{Pythia Contaminated} refers to p-values
when testing for membership of \emph{held-out} subsets
of datasets on a model contaminated 
with different subsets of the same datasets.
High p-values denote that our approach does not falsely 
detect membership.}
\label{tab:fpr_analysis_2}
\begin{center}
\begin{sc}
\begin{tabular}{lcc}
\toprule
    Dataset & \makecell{($\uparrow$) Pythia\\ \footnotesize{Uncontaminated}} & \makecell{($\uparrow$) Pythia\\
    \footnotesize{Contaminated}} \\
    \midrule
    TriviaQA                            &   0.52    &   0.28    \\
    ARC-C                               &   0.31    &   0.56    \\
    MMLU                                &   0.54    &   0.15    \\
    GSM8K                               &   0.38    &   0.47    \\
    Abstracts                           &   0.55    &   0.07    \\
    Blogs                               &   0.21    &   0.73    \\
\bottomrule
\end{tabular}
\end{sc}
\end{center}
\end{table}

\begin{table*}[th]
    \caption{ \textbf{Performance of models on the original datasets compared to the watermarked benchmarks.}
    We evaluate the models using the LM evaluation harness \cite{eval-harness} with the default settings,
    comparing performance on original benchmarks against two watermarking approaches:
    \textsc{Unicode}  substitutions \cite{wei2024proving} and \datasetname.
    We find that
    models obtain comparable performance on \datasetname-watermarked benchmarks,
    but crucially, \textbf{the relative ranking of LLMs remains unchanged across all benchmarks}, demonstrating the utility of watermarked benchmarks in comparing models.
    }
    \label{tab:benchmark_utility}
    \begin{center}
    \begin{sc}
    \begin{tabular}{ccccccccc}
    \toprule
    Dataset &
  Metric &
  Variant &
  \begin{tabular}[c]{@{}c@{}}Pythia \\ 1B\end{tabular} &
  \begin{tabular}[c]{@{}c@{}}Gemma-2 \\ 2B\end{tabular} &
  \begin{tabular}[c]{@{}c@{}}Mistral \\ 7B\end{tabular} &
  \begin{tabular}[c]{@{}c@{}}LLaMA-3 \\ 8B\end{tabular} &
  \begin{tabular}[c]{@{}c@{}}Gemma-2 \\ 9B\end{tabular} \\
  \midrule
\multirow{3}{*}{ARC-C}    & \multirow{3}{*}{0-shot} & Original          & 26.1  & 48    & 49.1  & 50.6  & 59.0 \\
                          &                         & Unicode           & 21.6  & 37.3  & 39.0  & 41.5  & 49.8 \\
                          &                         & \datasetnamespace & 26.3  & 46.8  & 49.1  & 50.5 & 57.1 \\
    \midrule
\multirow{3}{*}{MMLU}     & \multirow{3}{*}{5-shot} & Original          & 28.1 & 52.9  & 59   & 61.1 & 68.6 \\
                          &                         & Unicode           & 28.4 & 45.0  & 51.5 & 55.9 & 63.2 \\
                          &                         & \datasetnamespace & 28.8 & 51.6  & 56   & 61.8 & 68.4 \\
    \midrule
\multirow{3}{*}{TriviaQA} & \multirow{3}{*}{5-shot} & Original          & 12.4 & 52.7  & 67.2 & 68.9 & 70.1 \\
                          &                         & Unicode           & 1.1 & 23.6  & 46.0 & 44.3 & 54.8 \\
                          &                         & \datasetnamespace & 11.4 & 51.9  & 65.9 & 66.3 & 68.6 \\
    \midrule
\multirow{3}{*}{GSM8K}    & \multirow{3}{*}{5-shot} & Original          & 1.6  & 25.8  & 34.4 & 51.8 & 65.5 \\
                          &                         & Unicode           & 1.5  & 23.1  & 23.3 & 46.7 & 60.8 \\
                          &                         & \datasetnamespace & 2.2  & 27.2  & 37.5 & 54.9 & 65.8 \\
    \bottomrule
    \end{tabular}
    \end{sc}
    \end{center}
\end{table*}

\paragraph{False Positive Analysis.} \label{para:fpr}
To ensure the robustness of \datasetnamespace against false 
positives, we conduct two key experiments.
First, we apply our detection methodology to off-the-shelf
pretrained LLMs that have not been exposed to the watermarked benchmarks. 
The results for Pythia 1B, presented in the first 
column of Table \ref{tab:fpr_analysis_2}, show no false positives. 
We extend this analysis to models of different 
sizes and families in Table \ref{tab:fpr_analysis_all}, 
consistently finding no false positives across all tested models, 
confirming the robustness of \datasetnamespace against false positives.
Second, we perform a stronger test to evaluate whether \datasetnamespace 
detects the membership of the dataset
rather than just distributional differences due to different watermarking keys.
We create held-out subsets from the same benchmarks 
and watermark them using the identical public keys used for our contaminated versions. 
While these held-out sets share the same distribution 
and watermarking as our training data, they contain entirely different examples.
We then apply our detection methodology to test if these \textit{held-out} sets 
are falsely detected as members in our contaminated Pythia 1B model.
The second column of Table~\ref{tab:fpr_analysis_2}
shows consistently large p-values,
indicating \datasetnamespace successfully refutes membership for these \textit{held-out} sets.

\paragraph{Performance Without Watermarks Embedded.}

To validate our hypothesis that using a watermarked LLM to generate the rephrased copies 
of the benchmark enhances the statistical strength of our test,
we conduct experiments under the same settings as described above (\S\ref{para:setup}), 
but with rephrased copies generated without using a watermarked LLM.
The results, presented in Table \ref{table::main_results}, 
confirm that incorporating watermarked test sets 
significantly boosts the statistical power of our test, 
improving performance by at least two orders of magnitude across all benchmarks.

\subsection{Utility of Test Sets} \label{sec:utility_test_set}

Detecting contamination alone is insufficient; 
the watermarked content should retain the desired properties 
(for e.g., benchmarks should maintain their utility as reliable indicators of LLM performance). 
Using the lm-evaluation-harness framework \cite{eval-harness}, 
we assess five pre-trained LLMs on both original and watermarked benchmarks. 
Additionally, we measure semantic preservation using the P-SP metric \cite{wieting2021paraphrastic}.

\textbf{Our results}, presented in Table \ref{tab:benchmark_utility},
demonstrate that \datasetname-watermarked variants 
maintain benchmark utility: LLMs achieve similar absolute performance 
and the relative rankings of LLMs across all benchmarks are unaffected.
In contrast, \textsc{unicode} watermark \cite{wei2024proving} significantly degrades benchmark utility, 
with performance drops of up to $20$\% and does not preserve relative rankings. 
\datasetname-watermarked variants also result in
high 
semantic preservation (P-SP scores between $0.83$ \& $0.91$)
across all benchmarks. For reference,  the average score of human paraphrases is $0.76$ as per \cite{krishna2024paraphrasing}. 
These results are available in Table \ref{table::psp_scores}.

\begin{table}[t]
    \caption{\textbf{Semantic similarity scores} (P-SP) \cite{wieting2021paraphrastic}
    between original datasets and 
    their watermarked rephrases (higher is better).
    \textsc{Triv-QA} and \textsc{Abs.} refers to TriviQA and paper abstracts respectively.
    For reference: the P-SP value is $0.76$ 
    for human-written paraphrases as per a recent study~\cite{krishna2024paraphrasing}.}
    \begin{center}
    \footnotesize
    \begin{sc}
    \begin{tabular}{cccccc} %
    \toprule
     & Triv-QA & Arc-C & MMLU & GSM8K & Abs. \\
    \midrule
    P-SP   &  0.91  &  0.83  &  0.86  &  0.90 & 0.95  \\
    \bottomrule
    \end{tabular}
    \end{sc}
    \end{center}
    \label{table::psp_scores}
\end{table}

\begin{figure*}[ht]
    \centering
    \begin{subfigure}{0.48\textwidth}
        \centering
        \includegraphics[width=\linewidth]{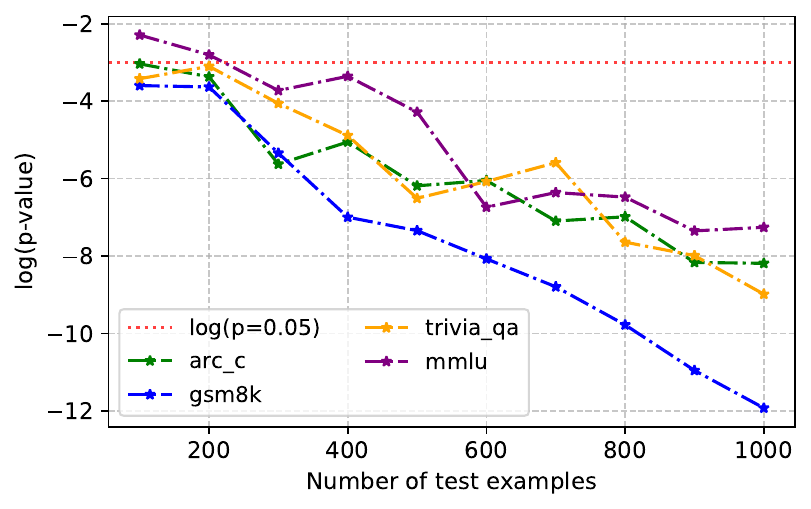}
        \caption{Log p-value vs sample size of benchmark ($n$)}
        \label{fig:ablation_num_samples}
    \end{subfigure}
    \hfill
    \begin{subfigure}{0.48\textwidth}
        \centering
        \includegraphics[width=\linewidth]{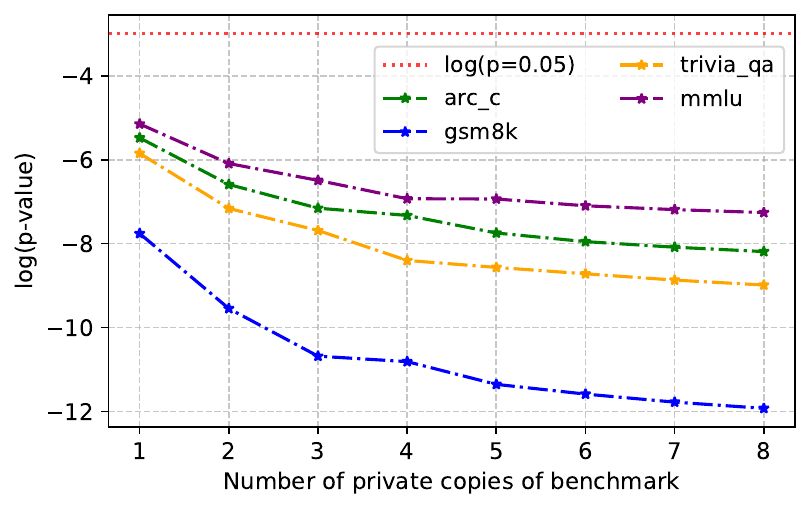}
        \caption{Log p-value vs private key count ($m$) }
        \label{fig:ablation_num_keys}
    \end{subfigure}
    \caption{
    \textbf{Impact of benchmark size ($n$) and private key count ($m$) on \datasetname's statistical power.}
    The dotted \textcolor{red}{red} line indicates the standard significance threshold ($p=0.05$).
    Lower values indicate stronger statistical evidence of contamination.
    }
    \label{fig:ablation_main_figure}
\end{figure*}

\subsection{Parameters Affecting the Power of the Test} \label{subsec:test_parameters}

\paragraph{Benchmark size.} 

To analyze the effect of sample size ($n$) on detection power,
we evaluate our test on 
benchmark subsets ranging from
$100$ to $1000$ examples.
For each size,
we average p-values across $10$ runs with different random seeds.
Our results, in Figure \ref{fig:ablation_num_samples}, demonstrate that our 
approach works even with just $600$ examples, 
where we consistently achieve low
p-values ($\approx 10^{-3}$) across all datasets.

\paragraph{Private key count.} 

Our proposed test compares the perplexity of the 
public version against the average perplexity
of $m$ private versions (Equation  \ref{eq:t_test_statistic_multi_avg}).  
Here we analyze how this hyperparameter ($m$)
affects the statistical power of our test.
As shown in Figure \ref{fig:ablation_num_keys},
increasing the number of private
keys strengthens detection up to a threshold of 5 keys,
beyond which we see negligible 
improvement.

\paragraph{Size of Pretraining Corpora.} \label{sec:exp_intentional_contam}

\begin{figure}[t]
    \begin{center}
    \includegraphics[width=0.94\columnwidth]{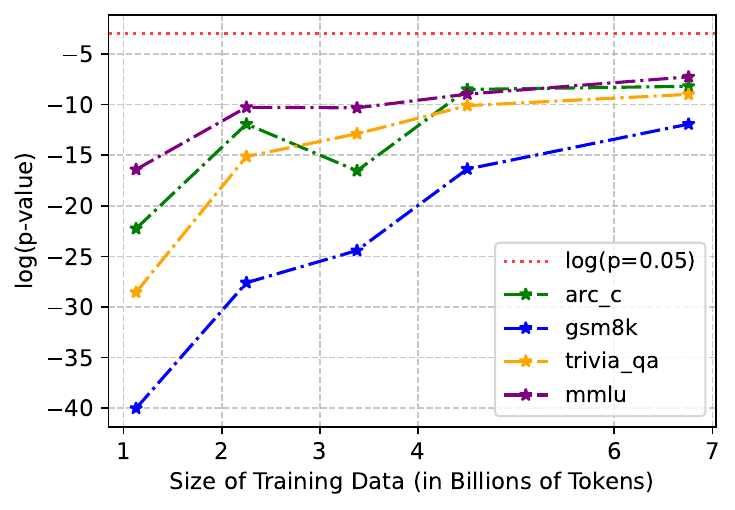}
    \caption{ \textbf{Log p-value vs pretraining corpus size.}
    We observe that the \emph{rate of decline} diminishes 
    as we increase the corpus size, with negligible
    drop between $4$B and $6$B.
    }
    \label{fig:ablation_corpora_size}
    \end{center}
\end{figure}

We analyze our test's  
effectiveness for different scales of pretraining data by combining
contaminated benchmarks with varying amounts of OpenWebText
data \cite{2023opencompass}.
We note that while the strength decreases with corpus size,
the \textit{rate of decline} diminishes substantially
beyond $4$ billion tokens, with minimal drop 
in detection strength between $4$ and $6$ billion tokens (Figure \ref{fig:ablation_corpora_size}).
Notably, these results are obtained with a modest 1B-parameter
model; given that larger models exhibit stronger memorization \cite{carlini2019secret}, 
we believe that \datasetnamespace
will detect membership for larger models.

\section{Real World Case Studies}

To demonstrate \datasetname's 
effectiveness in 
detecting \emph{unlicensed} use 
of copyrighted data in model training,
we present two expository case studies.
Specifically, we apply \datasetnamespace to
detect membership of
(1) abstracts from EMNLP 2024 proceedings \cite{emnlp-2024-main} and 
(2) articles from the AI Snake Oil newsletter \cite{arvidaisnakeoil}.

\paragraph{Paper Abstracts.} 
We sample $500$ papers from EMNLP 2024 proceedings \cite{emnlp-2024-main}
and generate watermarked rephrasings of their abstracts.
Additionally, we generate watermarked rephrasings 
for another set of $500$ abstracts, 
which we use as a \emph{held-out} validation 
set for our experiments.
The prompt templates used for rephrasing 
and examples of watermarked abstracts are provided 
in Appendix \ref{appendix:prompt_template} 
and Appendix \ref{appendix:rephrased_examples} , respectively.

To 
evaluate whether 
the semantic content 
of abstracts is preserved, 
we use the P-SP metric~\citep{wieting2021paraphrastic},
where watermarked 
abstracts 
achieve a high score of 0.95, indicating that 
the semantic content is largely preserved.
To further evaluate 
the acceptability 
of watermarked abstracts,
we conduct 
both an automated evaluation (using GPT-4) 
and a small-scale 
human study involving original authors.
In both evaluations, 
the 
original abstract and its watermarked rephrasing is compared, 
classifying the rephrased 
content into one of five options: 
\textit{preferred}, \textit{acceptable}, acceptable with \textit{minor revisions}, or \textit{major revisions}, and lastly \textit{unacceptable}.
Further details are available 
in Appendix \ref{sec:appendix_paper_abstracts}.

\begin{figure}[t]
    \begin{center}
    \includegraphics[width=0.94\columnwidth]{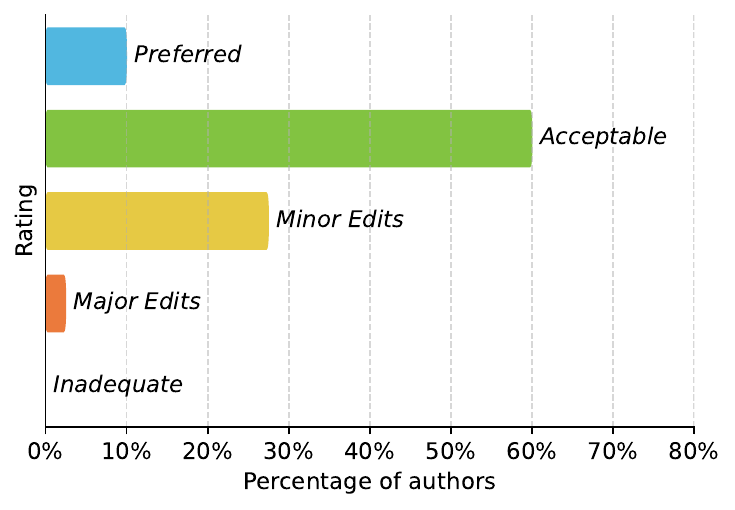}
    \caption{\textbf{Author evaluation of watermarked abstracts.}
    Out of the 40 total abstracts, for 4 of them the watermarked version was \textit{preferred}, for another 24 it was found to be \textit{acceptable},
    11 required \textit{minor edits} with just 1 requiring
    \textit{major edits}, suggesting that watermarking preserve quality.}
    \label{fig:human_study_plot}
    \end{center}
\end{figure}

For $1000$ watermarked abstracts, $99$\% were rated by GPT-4 
as either \textit{preferred} or \textit{acceptable}.
In a preliminary human study, 
we ask authors to review
rephrasings of their own abstracts.
Our results in Figure~\ref{fig:human_study_plot} show that
out of the $40$ watermarked abstracts evaluated,
authors find $24$ to be acceptable as is,
indicate $11$ could use minor edits,
and $4$ prefer the rephrased version over their self-written abstracts,
with just $1$ abstract requiring major edits.
Details of the evaluation 
are provided in Appendix \ref{prompt:gpt4_rephrase_quality}.

\paragraph{Blog Posts from AI Newsletter.}

We collect $56$ posts
from the popular AI Snake Oil newsletter \cite{arvidaisnakeoil},
and use $44$ for pretraining and hold $12$ for validation.
To demonstrate how \datasetnamespace 
could handle longer-form content, 
we adapt it to rephrase at the paragraph level, 
treating each paragraph as an 
independent datapoint for our test.
Note, while each paragraph serves as a datapoint
for our test, the blog posts are included 
in the pretraining corpora at the 
document level, following standard 
pretraining practices
(detailed in Appendix \ref{appendix:pretraining_details}).
We present the prompt used to rephrase
in Appendix \ref{appendix:prompt_template}

We evaluate \datasetname's ability 
to detect dataset membership by 
performing continual pretraining on the Pythia 1B model 
using a training corpus composed 
of watermarked paper abstracts ($\approx 105$K tokens), 
watermarked blog posts ($\approx 95$K tokens), 
and a subset of OpenWebText ($\approx 3.3$B tokens).
Additionally, to verify that \datasetnamespace can detect dataset membership 
for distinct datasets watermarked with the same key, 
we apply a consistent watermarking key
when generating the public versions of both datasets.

\paragraph{Results.}

Our results in Table \ref{table:emnlp_DI_results_main} demonstrate 
that \datasetnamespace 
effectively detects dataset membership 
for both paper abstracts and blog posts, 
achieving statistically significant p-values.
To compare, we evaluate LLM DI under 
different choices of validation set:
first, using private rephrases of the same documents 
and second, using a different held-out set of 
documents watermarked with the same public key. 
While LLM DI can detect membership for paper abstracts, 
it fails to do so for blog posts.
Further,  
membership inference attacks exhibit near-random performance (Table \ref{table::mia_metrics}).
To verify the robustness of \datasetnamespace against false positives,
we evaluate it under the two settings discussed earlier (\S\ref{para:fpr}).
Our results in Table \ref{tab:fpr_analysis_2} confirm that \datasetnamespace 
does not result in any false positives, reinforcing its reliability.

\begin{table}[t]
    \caption{ 
    \textbf{\textbf{Case studies.} }
    We report p-values of different approaches for detecting dataset membership (lower is better).
    LLM DI (same) uses the private rephrasing of the same documents, while
    LLM DI (different) uses different documents from a held out set from the same distribution.
    }
    \begin{center}
        \begin{sc}
            \begin{tabular}{lcc}
            \toprule
            Method $(\downarrow)$ & \makecell{Paper\\Abstracts} & \makecell{Blog\\Articles} \\
            \midrule
            LLM DI (same) & 0.15           & 0.44  \\
            LLM DI (different) & \textbf{0.05}  & 0.58   \\
            \midrule
            \makecell[l]{\datasetnamespace\\(w/o paired tests)}    & \textbf{0.01}     &   0.07            \\
            \datasetnamespace                       & \textbf{2.7e-12}  & \textbf{2.4e-3}   \\
            \bottomrule
            \end{tabular}
        \end{sc}
        \label{table:emnlp_DI_results_main}
    \end{center}
\end{table}

\section{Related Work} \label{sec:related_work}

We discuss the most closely related work below, focusing on
statistical approaches for detecting dataset membership, 
test-set contamination and the use of watermarks for detecting membership of datasets.
A more comprehensive review of related 
literature is provided in Appendix \ref{appendix:related_works}.

\paragraph{Dataset Membership.} 

A recent hypothesis-testing approach 
embeds random sequences in text or 
substitutes characters with visually-similar unicodes \cite{wei2024proving}. 
Similarly, 
\citet{meeus2024copyright} propose inserting ``copyright traps''
into documents to enhance document-level membership 
inference.
These methods then test the model’s preference for
these inserted sequences or substitutions.
However, such alterations impair machine readability, 
making them impractical for content creators.
Another recent proposal \cite{maini2024llmdatasetinferencedid}
selectively combines
membership inference attacks (MIAs) that provide positive signals 
for a given distribution and aggregates them 
to perform a statistical test on a dataset.
Their method assumes access to a validation set 
drawn from the same distribution as the 
target dataset and unseen by the model---a difficult requirement to satisfy.

A recent position paper~\citep{zhang2025membershipinferenceattacksprove} 
argues that methods attempting to estimate FPR by
collecting non-members a posteriori are statistically unsound,
a position that aligns with our analysis 
of PaCoST~\citep{zhang2024pacostpairedconfidencesignificance}.
We believe \datasetnamespace aligns with the criteria
for a sound membership proof presented in the paper.
Specifically, we use private members sampled from
the same distribution as the publicly released version $x$.
Since our private members are semantically equivalent 
to the public member, 
any causal effects of publishing $x$
would similarly affect the private members, 
ensuring the statistical validity of our approach.

\paragraph{Test Set Contamination.} 
While our focus is detecting membership of any arbitrary 
collection of documents, some recent \emph{statistical approaches} have 
focused on detecting test set contamination. 
A recent work \cite{oren2023provingtestsetcontamination} 
proposes a permutation test based 
on the canonical ordering in a benchmark but relies on the 
strong assumption of metadata contamination (canonical ordering of the dataset).
Another recent proposal \cite{zhang2024pacostpairedconfidencesignificance}
compares the model confidence on test instances and their rephrased counterparts.
However, as discussed earlier (\S\ref{sec:release_watermark_set}), 
LLMs may favor their own outputs, 
and this is an oft-overlooked confounder.
Additionally, there have been a few approaches
based on prompting models to reproduce near-exact test examples  
\cite{sainzchatGPTCheat, golchin2024timetravelllmstracing}.
However, the heuristic-y nature of these approaches 
prevents them from providing  
statistical evidence of contamination.

\paragraph{Watermarking for Dataset Membership.}
A few recent approaches have explored 
using LLM watermarks for membership detection.
Waterfall \citep{lau2024waterfallframeworkrobustscalable} proposes a 
watermarking scheme for protecting IP of text and further demonstrates 
how to detect unauthorized fine-tuning of LLMs on proprietary text data.
Specifically, to detect membership of a text, 
their approach prompts the target model with a prefix 
and detects the embedded watermarking 
in the generated new tokens to test for membership.
While effective in certain scenarios, 
their approach requires a higher level of memorization 
and has only been demonstrated in fine-tuning settings with multiple epochs. 
Additionally, their method is not applicable to domains like benchmarks 
where each sample is only a few tokens long. 
These limitations may restrict its practical utility for 
detecting membership of a dataset in the pretraining corpora of an LLM.

Another recent contemporaneous study ~\citep{sander2025detecting} 
proposes a similar approach where watermarks are embedded 
in benchmarks by reformulating the original questions 
with a watermarked LLM.
While employing a similar setup, 
their detection approach differs substantially from ours. 
Their method relies on detecting 
overfitting of the contaminated model
on token-level watermarking biases
to prove contamination, 
whereas our approach compares perplexity differences 
between the publicly released benchmarks and 
private versions watermarked with different keys.

\section{Conclusion \& Future Directions} \label{sec:discussion_with_conclusion}

In this work, we presented \datasetname, a 
statistical framework for detecting dataset membership, which can reliably
be used by content creators
to watermark their content, 
while preserving the utility, or 
the meaning, of the original content.
We demonstrated \datasetname's effectiveness 
in detecting test-set contamination
through comprehensive experiments. 
Our ablation studies systematically analyzed 
how detection strength varies with dataset size, 
the number of private versions, and pretraining corpus size. 
We validated the real-world  applicability of our approach
through two case studies: detecting paper abstracts
and blog posts in pretraining data.

There are several \textbf{important limitations} of our work: first, watermarks must be embedded
before the content is released online,
making it inapplicable to 
already published content.
We believe this is a fundamental limitation 
shared by existing statistical methods, 
as they require knowledge of the data-generating process to construct a valid null distribution.
Second, our method requires 
access to token probabilities 
from the model (gray box access). 
Third, while our human study showed that 
majority of authors found the rephrasings to be acceptable,
rephrasing could introduce errors in the content. 
However, we believe this will be less of a concern moving forward
as general model capabilities, 
including paraphrasing quality, continue to improve.
Finally, due to computational constraints, 
we evaluated our approach using continual pretraining 
rather than training models from scratch.
While our results demonstrate effectiveness 
in this setting outperforming baselines, 
future work could validate these findings 
using models that are trained from scratch.

Future work could 
explore the optimal watermarking 
strength 
for different data distributions (and use cases) 
to balance 
a (plausible)
trade-off between 
detectability and 
quality of watermarked content.
Future work could also validate, or 
extend, our approach 
to other domains, 
such as code, speech, images or videos.

\section*{Acknowledgments}

We thank the reviewers for their feedback.
We sincerely thank all participants 
in our evaluation study
for their valuable time.
This work was supported in part by the AI2050 program at Schmidt Sciences (Grant G-24-66186). Additionally, DP is grateful to Adobe Inc., Pratiksha Trust and the National Payments Corporation of India (NPCI) for generously supporting his group’s research.

\section*{Impact Statement}

Our work studies the problem of detecting 
unauthorized usage of data for model training.
In the current landscape, where model developers
are reluctant to share details about their pretraining corpora, 
we believe our proposal holds potential to
considerably increase transparency in model
training. Our tool could be beneficial 
to content creators seeking to
protect their work from unauthorized use.

Additionally, our work has implications for
the broader AI ecosystem. 
By detecting test-set contamination, 
our approach can help researchers obtain
more accurate estimates 
of model capabilities and track AI progress.

\bibliography{refs}
\bibliographystyle{icml2025}

\newpage

\appendix
\onecolumn

\section{Additional Results}

\begin{table}[ht]
\caption{Size of evaluation benchmark used in the
intentional contamination experiment (\S\ref{sec:exp_intentional_contam}). 
Each benchmark is subsampled to 1,000 examples, 
with each injected benchmark making up less 
than 0.001\% of the entire pretraining corpus,
which consists of $6.7$ billion tokens.
Each benchmark is injected exactly once into the corpus without any duplication.
}
\label{tab:contamination_sizes}
\begin{center}
\begin{sc}
\begin{tabular}{lcc}
\toprule
    \textsc{Benchmark}     &  \textsc{Size (Tokens)} & \textsc{\makecell{\% Pretraining Data}}  \\
    \midrule
    \textsc{TriviaQA}      &  34609 & 5.1e-4     \\
    \textsc{Arc-C}         &  36863 & 5.5e-4     \\
    \textsc{MMLU}          &  42548 & 6.3e-4     \\
    \textsc{GSM8k}         &  61132 & 9.0e-4     \\
\bottomrule
\end{tabular}
\end{sc}
\end{center}
\end{table}

\begin{table*}[ht]
    \caption{\textbf{Comparison of Membership Inference Attacks (MIA) performance across different datasets.}
    We report AUC scores for three MIA methods under two settings:
    \textit{Same Documents:} public rephrases in training vs private rephrases of the same documents, and 
    \textit{Different Documents:} public rephrases in training vs different unseen documents from the same dataset.
    AUROC score of $\approx 0.5$ indicates performance no
    better than random guessing.
    }
    \begin{center}
    \begin{sc}
    \resizebox{\textwidth}{!}{
    \begin{tabular}{lcccccccccc}
    \toprule
    & \multicolumn{5}{c}{ \textit{Same Documents.} ($\uparrow$) } & \multicolumn{5}{c}{ \textit{Different Documents.} ($\uparrow$) } \\
    \cmidrule(lr){2-6} \cmidrule(lr){7-11}
    Dataset         & PPL       & Zlib  & Min-K & Min-K++ & DC-PDD  & PPL  &   Zlib &   Min-K   & Min-K++   & DC-PDD  \\
    \midrule
    TriviaQA        &  0.50     & 0.50  & 0.50  &  0.50   & 0.52    & 0.46 &   0.57 &   0.48 & 0.44 & 0.58      \\
    Arc-C           &  0.50     & 0.50  & 0.50  &  0.49   & 0.51    & 0.49 &   0.50 &   0.48 & 0.45 & 0.52      \\
    MMLU            &  0.48     & 0.49  & 0.48  &  0.49   & 0.52    & 0.43 &   0.48 &   0.44 & 0.45 & 0.52      \\
    GSM8k           &  0.50     & 0.50  & 0.50  &  0.50   & 0.52    & 0.47 &   0.47 &   0.48 & 0.48 & 0.52      \\
    Paper Abstracts &  0.48     & 0.49  & 0.48  &  0.46   & 0.53    & 0.41 &   0.46 &   0.42 & 0.40 & 0.55      \\ 
    Blog Articles   &  0.50     & 0.51  & 0.51  &  0.49   & 0.50     & 0.49 &   0.51 &   0.48 & 0.46 & 0.51      \\
    \bottomrule
    \end{tabular}
    }
    \end{sc}
    \end{center}
    \label{table::mia_metrics}
\end{table*}

\begin{table*}[h]
\caption{ \textbf{False positive analysis on off-the-shelf LLMs.}
We apply \datasetnamespace on LLMs
that have not seen the datasets and 
report the p-values.
Our results (high p-values) show that our method
is robust against false positives.
}
\label{tab:fpr_analysis_all}
\begin{center}
\begin{sc}
\begin{tabular}{lccccc}
\toprule
    {Dataset ($\uparrow$)}  &  
    \begin{tabular}[c]{@{}c@{}}Pythia \\ 1B\end{tabular} &
    \begin{tabular}[c]{@{}c@{}}Gemma-2 \\ 2B\end{tabular} &
    \begin{tabular}[c]{@{}c@{}}Mistral \\ 7B\end{tabular} &
    \begin{tabular}[c]{@{}c@{}}LLaMA-3 \\ 8B\end{tabular} &
    \begin{tabular}[c]{@{}c@{}}Gemma-2 \\ 9B\end{tabular} \\
    \midrule
    TriviaQA    & 0.52  & 0.91 & 0.94   & 0.65  & 0.91  \\
    ARC-C       & 0.31  & 0.25 & 0.12   & 0.26  & 0.37  \\
    MMLU        & 0.54  & 0.41 & 0.29   & 0.24  & 0.43  \\
    GSM8k       & 0.38  & 0.16 & 0.26   & 0.71  & 0.37  \\
    {Paper Abstracts}    & 0.55  & 0.74  & 0.83  & 0.63  & 0.89  \\
    {Blog Articles}      & 0.21  & 0.72  & 0.74  & 0.88  & 0.12 \\
\bottomrule
\end{tabular}
\end{sc}
\end{center}
\end{table*}

\subsection{Detecting Partial Contamination}

\begin{figure}[h]
    \begin{center}
    \includegraphics[width=0.75\columnwidth]{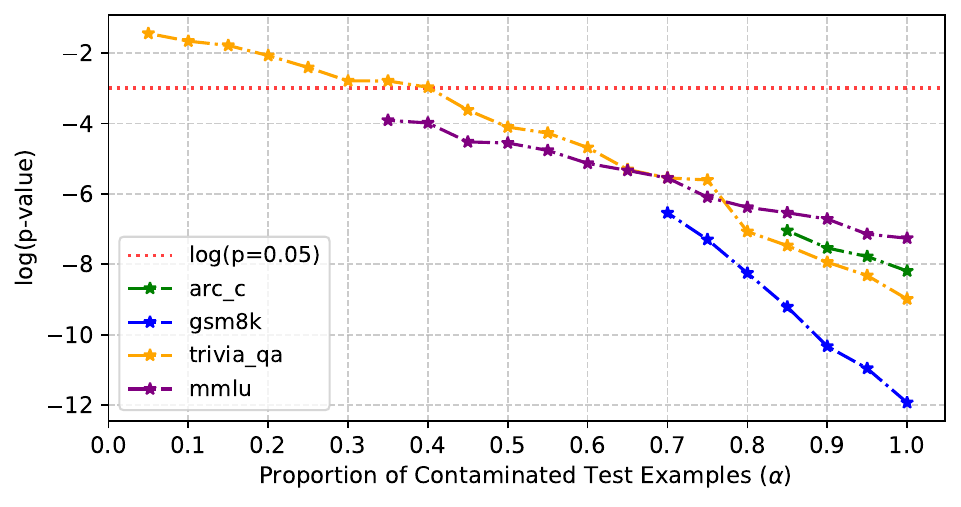}
    \caption{\textbf{Log p-value vs proportion of benchmark that is contaminated}.
    We plot the log p-value against the proportion of test examples that are leaked
    to analyze the sensitivity of our test to detect contaminated in scenarios where
    the benchmark is only partially contaminated ({lower is better}).}
    \label{fig:appendix_ratio_benchmark}
    \end{center}
\end{figure}

In practice, benchmarks may be partially contaminated, 
where only a subset of test examples appears in the pretraining corpora. Understanding the impact of partial contamination 
is critical because benchmark owners 
cannot identify which specific 
test examples have been leaked.
This study complements our earlier analysis 
in Section \ref{sec:exp_intentional_contam}, 
by focusing on the sensitivity of 
our approach under varying proportions ($\alpha$)
of contaminated examples within a fixed benchmark size ($n$).

Our results in Figure \ref{fig:appendix_ratio_benchmark} highlight that 
as $\alpha$ increases the detection strength improves, with p-values
dropping below $10^{-3}$ when majority of the benchmark is contaminated.
We also observe that \datasetnamespace reliably detects contaminated 
even when only 40\% of the test examples are contaminated.
Our findings confirm that \datasetnamespace successfully 
identifies contamination with high statistical significance, 
even in scenarios of partial contamination.

\newpage

\section{P-SP Metric} \label{appendix:psp_metric}

To validate semantic preservation 
in our watermarking process, we employ P-SP \cite{wieting2021paraphrastic}, 
a state-of-the-art semantic similarity model. 
P-SP uses embedding averaging trained on a 
large corpus of filtered paraphrase data, and 
has been shown to effectively distinguish 
between true paraphrases and unrelated text. 
As evidenced by \citet{krishna2024paraphrasing}, 
P-SP assigns an average score of 0.76 to human-created paraphrases 
in the PAR3 dataset \cite{thai2022exploring}, 
while random paragraph pairs from the same book score only 0.09. 
Table \ref{table::psp_scores} reports the average 
P-SP scores between original benchmarks 
and their watermarked versions across 9 random \emph{hash keys}.
Our watermarked versions achieve high P-SP scores (0.83-0.95) 
across all benchmarks, substantially exceeding 
the average score for human paraphrases, 
indicating strong semantic preservation.

\section{Bag-of-Words Classifier} \label{appendix:bagofwords}

We train a random forest classifier on the 
\emph{bag-of-words} feature representations 
for the datasets. 
The classifier is trained on 80\% of the member and non-member sets, 
with evaluation performed on the remaining 20\%. 
Results are aggregated over a 5-fold cross-validation.
The detailed results are presented in Table \ref{table::iid_check}.

\begin{table}[ht]
    \caption{AUROC using \textit{bag-of-words} features
    to distinguish between different versions of datasets.
    The first column shows AUROC for distinguishing original datasets 
    from their rephrased versions, 
    where high values ($>0.8$) indicate clear 
    distributional differences.
    The second column shows AUROC 
    for distinguishing between \emph{public} and 
    \emph{private} watermarked versions, 
    where values near $0.5$ indicate 
    distributional similarity.}
    \begin{center}
    \begin{sc}
    \begin{tabular}{lcc}
    \toprule
    Dataset     & \makecell{Original vs \\Rephrased}  & \makecell{Public vs \\Private}  \\
    \midrule
    TriviaQA            &  0.66  &  0.51  \\
    Arc-C               &  0.83  &  0.52  \\
    MMLU                &  0.83  &  0.53  \\
    GSM8k               &  0.84  &  0.57  \\
    paper abstracts     &  0.86  &  0.57  \\
    \bottomrule
    \end{tabular}
    \end{sc}
    \end{center}
    \label{table::iid_check}
\end{table}

\section{Perplexity} \label{appendix:perplexity}

Perplexity (\emph{PPL}) measures how well a language model 
predicts a given text sequence $S$,
with lower values indicating better prediction. 
For an auto-regressive language model $\theta$ and text sequence $S$,
tokenized as a sequence of N tokens $\{s_1,\ldots,s_N\}$,
perplexity is computed as the exponent of the loss.
Formally:

\begin{equation}
\begin{aligned}
\label{eq:perplexity_def}
{\textit{PPL}}_{\theta}(S) & = \exp\left(\mathcal{L}_{ \theta }(S)\right)
\end{aligned} 
\end{equation}

Where the loss $\mathcal{L}_{\theta}$ is defined as: 

\begin{equation}
\begin{aligned}
\label{eq:model_loss}
\mathcal{L}_{\theta}(S) & = -\frac{1}{N}\sum_{i=1}^{N} \log\left( \mathcal{P}_{\theta}(s_i | s_{<i}) \right)
\end{aligned} 
\end{equation}

Here 
$\mathcal{P}_{\theta}(s_i | s_{<i})$
denotes the predicted probability for token $s_i$ 
by the language model $\theta$ given the context of previous tokens $\{s_1,\ldots,s_{i-1}\}$.

\section{Pretraining Details} \label{appendix:pretraining_details}

We continually pretrain Pythia 1B on 
intentionally contaminated OpenWebText 
Test case instances from the benchmark were randomly 
inserted between documents from OpenWebText. 
We trained for 1 epoch of 46000 steps with an effective batch 
size of 144 sequences and sequence length of 1024 tokens.
We used the AdamW optimizer \cite{loshchilov2019decoupledweightdecayregularization} with a learning rate of $10^{-4}$, $(\beta_1, \beta_2) = (0.9, 0.999)$ and no weight decay.

\section{Watermark for Large Language Models} \label{appendix:watermark_llms}

In work, we use the prominent 
KGW \cite{pmlr-v202-kirchenbauer23a} watermarking scheme.
KGW scheme uses a hash function that takes
the context (preceding tokens) and a hash key $h$ to partition
the vocabulary $V$ into two disjoint sets at each generation
step: a green list $G$ and a red list $R$.
Formally, for a language model $\mathcal{M}$ 
with vocabulary $V$, 
and a prefix comprising tokens $\mathbf{w}_1, \mathbf{w}_2, \ldots, \mathbf{w}_n$, 
the scheme involves first computing the logits
 $\mathcal{M}(\mathbf{w}_1 \ldots, \mathbf{w}_n) = (l_1, \ldots, l_{|V|})$ 
 of the language model that would ordinarily be used to predict the 
subsequent token. 
As per a hyper-parameter $k$, the last $k$ tokens,   
$\mathbf{w}_{n-k+1}$ to $\mathbf{w}_{n}$,
are then fed to a pseudo-random function $F$
to partition $V$ into a green list $G$
 and a red list  $R$
such that $|G| + |R| = |V|$.
Finally, the logits 
corresponding to the tokens in the green list, $G$, are boosted
by $\delta$ ($\delta > 0$). 
Specifically, in our work we set $k = 1$ and $\delta = 1.0$ as the chose hyperparameters.
The watermark can then be detected through a one-proportion z-test 
on the fraction of green tokens in the generated text.

\section{Related Works} \label{appendix:related_works}

Our works relates to a large literature of work on 
membership inference (\S\ref{subsec:appendix_mia}),
dataset membership (\S\ref{subsec:appendix_di})
and test-set contamination detection (\S\ref{subsec:appendix_contamination})
in large language models.

\subsection{Watermarking for Dataset Membership} \label{subsec:appendix_wtm_di}

\citet{liu2023watermarking} propose TextMarker,
a backdoor-based membership inference technique 
for protection of \textit{classification datasets}.
Their approach watermarks each original sample $(x,y)$ 
by inserting specific triggers (character or word-level substitutions) into $x$, 
creating a backdoored sample $x_t$. 
They then assign an altered target label $y_t$ $(y_t \neq y)$ to this modified input. 
Membership detection is performed by testing whether 
a model $f$ produces the watermarked label with high probability: $\Pr(f(x_t) = y_t)$.
While effective, TextMarker is specifically designed 
for classification datasets and it is not trivial 
to extend it to other kinds of benchmarks or the broader problem 
of dataset membership detection, limiting its real-world applicability.

Recent work has proposed detecting 
watermark signals in suspect model outputs to determine dataset membership. 
Waterfall \citep{lau2024waterfallframeworkrobustscalable} enables creators 
to watermark their text using a robust watermarking scheme 
that leverages an LLM for rephrasing. 
They further demonstrate that watermarks in fine-tuning data, 
persist in downstream LLM outputs, allowing for membership detection. 
While both approaches use LLM-based rephrasing, 
Waterfall relies on detecting watermarks directly in model generations, 
whereas our method uses multiple 
watermarked versions to detect perplexity divergences.
While Waterfall is effective for text watermarking, 
it has limited utility for pretraining membership detection. 
It relies on strong overfitting and presumes 
that the model is trained on datasets over multiple epochs. 
Additionally, their approach is not applicable 
to short text segments commonly found in benchmarks.
Another recent contemporaneous study ~\citep{sander2025detecting} 
proposes a similar approach where watermarks are embedded 
in benchmarks by reformulating the original questions with a watermarked LLM.
While employing a similar setup, 
their detection approach differs substantially from ours. 
Their method relies on detecting model overfitting 
on the green tokens in the watermarked benchmark to prove contamination, 
whereas our approach compares perplexity differences 
between the publicly released benchmarks and 
private versions watermarked with different keys.

\subsection{Membership Inference} \label{subsec:appendix_mia}

Membership inference, initially proposed by \citet{shokri2017membershipinferenceattacksmachine}, 
is a long-standing problem in machine learning: 
given a data point and a machine learning model, 
determine whether that data point was used to train the model.
MIAs for LLMs are broadly based on applying 
pre-defined thresholds to membership scores that are typically
based on \emph{loss-based} metrics.
We briefly describe the specific membership scores proposed by 
different MIAs that we employ in our experiments.

\begin{itemize}
    \item \textsc{Perplexity:}  Proposed by \citet{yeom2018privacy}, this MIA uses loss (perplexity in the context of LMs) as 
    the scoring metric. However, this approach suffers from high false positives  as it tends to classify naturally predictable sequences as members of the training set.
    \item \textsc{zlib Entropy} \citep{carlini2021extracting}  computes a score by taking the ratio between the model's perplexity and the zlib compression size of the text. Lower ratios indicate potential membership in the training data.
    \item \textsc{Min-K\%} \cite{shi2024detectingpretrainingdatalarge} computes the score by averaging the probabilities of the $k\%$ least likely tokens in a sequence. By focusing on the least likely tokens, it aims to solve the false positive problem with perplexity.
    \item  \textsc{Min-K\%++} \cite{zhang2024min} compares the probability of the target token with the expected probability of all tokens within the vocabulary. It is based on the insight that each training token will tend to have higher probability relative to many other candidate tokens in the vocabulary.
    \item \textsc{DC-PPD} \cite{zhang2024pretrainingdatadetectionlarge} computes the divergence between the token probability distribution and the token frequency distribution for detection. 
\end{itemize}

\subsection{Detecting Dataset Membership} \label{subsec:appendix_di}

Detecting dataset membership addresses the challenge of
detecting whether a given dataset was used by 
LLM developers in pretraining. 
Unlike membership inference attacks (MIAs), 
which focus on identifying whether individual sequences 
were included in a model's training data, 
dataset membership concerns verifying 
the inclusion of a collection of documents.

\citet{wei2024proving} propose a hypothesis-testing approach 
to detect membership by inserting random sequences or 
Unicode character substitutions as data watermarks. 
This method works by testing the model's preference 
for the inserted data watermarks against other random data watermarks.
First, their proposed watermarks can 
impact machine readability, 
affecting search engine indexing and 
retrieval-augmented generation (RAG) pipelines. 
More critically, unicode substitutions can significantly
alter tokenization processes, potentially compromising 
the utility of evaluation benchmarks. 
Although these limitations may be manageable for some creators, 
Our approach offers an alternative 
that better preserves content quality
while maintaining detection capability.
Another recent proposal \cite{maini2024llmdatasetinferencedid}
is to selectively combining MIAs
that provide positive signal for a given distribution, 
and aggregating them to perform a statistical test on a given dataset.
Their method assumes access to a \emph{validation} set
drawn from the same distribution as 
the target dataset and unseen by
the model--a requirement that can be challenging to satisfy
in many practical scenarios.

\citet{meeus2024copyright} propose inserting “copyright traps” 
into documents to enhance document-level membership 
inference for smaller models that lack natural memorization.
\citet{liu2023watermarking} introduce a backdoor-based dataset inference approach. 
However, these methods rely on heuristics and 
do not provide the false positive guarantees that hypothesis-testing-based approaches offer.

Recent studies \cite{maini2024llmdatasetinferencedid, duan2024membership, das2024blindbaselinesbeatmembership, Meeus2024SoKMI} suggest 
that detecting sequence level membership in LLMs 
trained on trillions of tokens in a single epoch is likely infeasible. 
These studies also highlight the limited efficacy of MIAs for LLMs, 
showing that such approaches barely outperform random guessing. 
Moreover, the apparent success of MIAs in certain scenarios 
can often be attributed to distributional differences 
between the \textit{member} and \textit{non-member} sets used in evaluations, 
rather than their ability to reliably infer true membership.

\subsection{Test Set Contamination Detection} \label{subsec:appendix_contamination}

There have been 
a few recent third-party approaches that are focused on 
detecting test-set contamination in LLMs. 
Heuristic prompting-based methods \cite{sainzchatGPTCheat, golchin2024timetravelllmstracing} attempt to 
detect contamination by prompting models to
reproduce exact or near-exact test examples. 
Reproducing verbatim examples requires a high 
level of memorization which typically requires a 
high duplication of test examples \cite{carlini2021extracting} 
and strong memorization capabilities 
typically absent in smaller models \cite{meeus2024copyright}.
The heuristic nature of these approaches 
prevents them from providing a
statistical evidence of contamination.

Statistical approaches to detect contamination are limited.
\citet{oren2023provingtestsetcontamination} build on the principle 
that in absence of data contamination,
all orderings of an \emph{exchangeable} 
test set should be equally likely.
Their work relies on the strong assumption 
of metadata contamination (canonical ordering of the dataset)--a 
presumption that can often be violated.
Another recent proposal \cite{zhang2024pacostpairedconfidencesignificance}
uses a statistical test to compare model confidence 
on original test instances and 
their rephrased counterparts.
However, as discussed earlier,
their null hypothesis can be invalid
due to LLMs' inherent bias towards
machine-generated content.

\section{Radioactivity of Watermarks} \label{sec:appendix_radioactive_wtms}

\citet{sander2024watermarking} proposed methods
to detect when watermarked texts 
are used as fine-tuning data for an LLM.
Their approach is based on the insight 
that training on  watermarked texts 
leaves detectable traces of the watermark signal
in the resulting model due to 
token-level overfitting.
In a recent contemporaneous study,
\citet{sander2025detecting} extended this
approach to detect benchmark contamination.
Specifically, they propose
watermarking benchmarks before release
and later detecting traces 
left by the watermarked benchmarks
through a statistical test.
Since the statistical test relies on
token-level overfitting,
their approach requires duplication 
and stronger watermarks,
which introduce more distortion into the rephrasings.
Additionally, the tokenizer-dependent nature of 
detecting watermarks limits the applicability 
of their approach, as the rephrasing model 
and contaminated model need to share the same tokenizer.

Given the requirement that the rephrasing 
and contaminated LLM 
should share the same tokenizer,
we conduct additional controlled 
experiments comparing the approaches.
We rephrase with Llama-3.1-8B Instruct~\citep{llama3modelcard}
with top-p sampling with $p=0.7$ and temperature $=0.5$,
matching the sampling parameters used in the original study.
For watermarking, we use KGW~\citep{kirchenbauer2024reliabilitywatermarkslargelanguage} scheme,
with context window of size $2$,
split ratio ($\gamma$) of $0.5$ \&
and boosting value ($\delta$) of $2$.
We create a contaminated corpus of $2$ billion tokens
following our methodology in 
Section \ref{subsec:exp_intentional_contam},
but with a duplication count of 4, 
meaning each benchmark sample is inserted 
four times in the pretraining corpora.

We compare \datasetnamespace with~\citet{sander2025detecting} 
for detecting benchmark contamination in Table~\ref{table::appendix_radioactive_watermarks}.
With a moderate watermarking strength ($\delta=2.0$)
and repetition count of $4$,
the radioactivity based approach
fails to detect contamination 
of watermarked test examples,
while \datasetnamespace achieves significantly low p-values.
These results align with the original paper's findings,
which indicated that their method requires
around 16 repetitions to achieve low p-values.

\begin{table*}[th]
\caption{\textbf{P-values for detecting \emph{test-set contamination}.}
We compare our proposed \datasetnamespace approach with 
detection based on radioactivity~\cite{sander2025detecting}.
Rows marked \textbf{0} denote vanishingly small p-values.
Across both the benchmarks, 
\datasetnamespace consistently 
achieves lower p-values  (lower is better).}
\begin{center}
    \begin{center}
    \begin{sc}
    \begin{tabular}{ccc}
    \toprule
    & \multicolumn{2}{c}{Benchmark ($\downarrow$)} \\
    \cmidrule(lr){2-3}
    Method           & Arc-C & MMLU \\
    \midrule
    \makecell{Radioactivity \\~\citep{sander2025detecting}} & 0.61        &   0.13       \\
    \makecell{\datasetnamespace}                            &  \textbf{0} &  \textbf{0}  \\
    \bottomrule
    \end{tabular}
    \end{sc}
    \end{center}
\end{center}
\label{table::appendix_radioactive_watermarks}
\end{table*}

\section{Case Study: Detecting Research Paper Abstracts in Pretraining Data} \label{sec:appendix_paper_abstracts}

To demonstrate the broader applicability of \datasetnamespace for 
detecting dataset membership across different forms of content,
we explore its effectiveness in detecting membership of 
abstracts of papers from EMNLP '24 proceedings \cite{emnlp-2024-main}.
We evaluate both the preservation of 
academic writing quality in watermarked abstracts 
and the effectiveness of \datasetnamespace in detecting 
their inclusion in training data.

\paragraph{Experimental Setup.}

We sample $500$ papers from EMNLP 2024 proceedings
and create watermarked versions of their abstracts 
following our methodology from Section \ref{sec:methodology}.
The prompt template and examples of rephrased abstracts are presented
in Appendix \ref{prompt:llama_prompt_emnlp} 
and Appendix \ref{appendix:rephrased_examples_abstracts} respectively.
To evaluate detection capability,
we perform controlled experiments 
on the Pythia 1B model \citep{biderman2023pythiasuiteanalyzinglarge}
through continual pretraining.
The pretraining corpora consists of a mixture of 
the public watermarked versions of these abstracts and 
a subset of OpenWebText (approximately $3$ billion tokens).
The abstracts comprise approximately $100$K tokens, 
representing just $0.003$\% of the pretraining corpus.

\paragraph{Results.}

Table \ref{table:emnlp_DI_results_appendix} demonstrates \datasetname's effectiveness 
in detecting dataset membership. 
Our approach achieves a near-zero p-value 
($ \approx 10^{-12}$),
indicating strong statistical evidence of membership.
For comparison, LLM DI \cite{maini2024llmdatasetinferencedid}
achieves a p-value of $0.05$ with access to a 
\emph{validation} set of unseen abstracts from the same conference
and is unable to detect membership using the 
privately held counterparts of the same abstracts included 
in the pretraining data as the \emph{validation} set.
In Table \ref{table::mia_metrics} we evaluate
state-of-the-art MIAs and finding that they  
perform no better than random chance (AUROC $\approx$ 0.5).
Our findings corroborate with recent studies 
\cite{duan2024membership, maini2024llmdatasetinferencedid, das2024blindbaselinesbeatmembership}
that highlight the failure of sequence level MIAs on LLMs.

\paragraph{Quality Evaluation.} 

To evaluate the quality of watermarked abstracts, 
we use GPT-4 \cite{openai2024gpt4technicalreport} as a judge following the 
prompt template in Figure \ref{prompt:gpt4_rephrase_quality}.
Each abstract was classified into one of five quality tiers. 
Our analysis shows that $82.7$\% of the watermarked abstracts were rated as \textit{preferred} 
and $16.3$\% as \textit{acceptable} indicating that $99$\% maintain high academic quality. 
Only $1$\% required \textit{minor revisions},
with none requiring \textit{major revisions} or deemed \textit{inadequate}.

Since LLMs often exhibit systematic preferences 
for LLM-generated text over human-written 
content~\citep{liu2023gevalnlgevaluationusing, Mishra2023CharacterizingLL, Laurito2024AIAB}, 
we additionally conduct 
a human study involving
the original authors. We asked $40$ authors to rate
watermarked versions of their own abstracts using the same quality tiers.
The human evaluation strongly corroborates our
automatic assessment, 
with most watermarked versions being \textit{preferred} 
or \textit{acceptable}: 
$4$ authors \textit{preferred} the watermarked version, 
$24$ authors rated the watermarked abstracts as \textit{acceptable},
$11$ indicated the text required \textit{minor revisions}
and just $1$ indicating that their rephrased abstract requires major edits.

Additionally, we measure semantic preservation 
using the P-SP metric \cite{wieting2021paraphrastic}, 
finding an {average score of $\textbf{0.95}$}
between original and watermarked abstracts, 
demonstrating strong semantic similarity.

\begin{table}[h]
    \caption{ 
    \textbf{Comparison of different approaches for 
    detecting membership of paper abstracts.}
    \textbf{Bold} indicates statistically significant results ($p < 0.05$).
    Our approach results in lower p-values compared 
    to other approaches (lower is better).}
    \begin{center}
        \begin{sc}
            \begin{tabular}{lc}
            \toprule
            Method & P-value ($\downarrow$) \\
            \midrule
            LLM DI \cite{maini2024llmdatasetinferencedid} (1) & 0.15  \\
            LLM DI \cite{maini2024llmdatasetinferencedid} (2) & \textbf{0.05}  \\
            \midrule
            \datasetnamespace (w/o paired tests)    & \textbf{0.01}    \\
            \datasetnamespace                       & \textbf{2.7e-12} \\
            \bottomrule
            \end{tabular}
        \end{sc}
        \label{table:emnlp_DI_results_appendix}
    \end{center}
\end{table}

\begin{tcolorbox}[colback=blue!8, colframe=gray!80, rounded corners, sharp corners=northeast, sharp corners=southwest,title=Prompt Template to Evaluate Quality of the Rephrased Abstracts using GPT4,label=prompt:gpt4_rephrase_quality]

\texttt{You will be given an original abstract and its rephrased version. Your task is to evaluate the quality of abstract rewrites for ML research paper based on:\\ \\1. \textbf{Meaning Preservation}\\2. \textbf{Clarity}\\3. 
\textbf{Technical Accuracy}\\ \\ Evaluate the rewritten abstract and assign one of these ratings:\\- \textbf{Preferred:} The rewrite improves upon the original in terms of clarity and readability while maintaining full technical accuracy.\\- \textbf{Acceptable:} The rewrite matches the original in quality and could serve as a direct replacement without requiring changes.\\- \textbf{Minor Revisions:} The rewrite is promising but requires minor edits to reach the original's quality.\\- \textbf{Major Revisions:} The rewrite has significant issues with meaning preservation, clarity, or technical accuracy and requires major edits.\\- \textbf{Inadequate:} The rewrite fails to convey the original research effectively due to critical flaws in meaning, clarity, or technical accuracy.\\\\Here are the abstracts:\\\\Original Abstract: \{original\_abstract\} \\Rephrased Abstract: \{watermarked\_abstract\}\\\\Provide a short explanation of your rating, followed by your final rating in the format:\\ \textbf{Final Rating:} \{rating\} }

\end{tcolorbox}

\section{Case Study: Detecting ML Blog Posts in Pretraining Data} \label{sec:appendix_blog_posts}

The inclusion of copyrighted material in LLM training data has
emerged as a significant concern, leading to legal disputes,
such as the lawsuit between New York Times and OpenAI \cite{nytimes}, among others.
Through a case study, we demonstrate how \datasetnamespace can help creators detect 
potential unauthorized use of 
their content in model training.
Specifically, we use \datasetnamespace to detect the membership
of the popular AI Snake Oil newsletter \cite{arvidaisnakeoil}.

\paragraph{Experimental Setup.} 

We collect $56$ blogs from the newsletter,
creating watermarked versions of each newsletter using, 
the prompt template is presented in Figure \ref{prompt:llama_prompt_blog}.
We randomly select a subset of $44$ blogs that we include in 
pretraining corpora and keep the remaining $12$ blogs as a
\emph{validation} set that is unseen by the model.
To evaluate detection capability,
we perform controlled experiments 
on the Pythia 1B model \citep{biderman2023pythiasuiteanalyzinglarge}
through continual pretraining.
The pretraining corpora consists of a mixture of 
the public watermarked versions of these abstracts and 
a subset of OpenWebText (approximately $3$ billion tokens).
The abstracts comprise approximately $94$K tokens, 
representing just $0.003$\% of the pretraining corpus.

\begin{table}[ht]
    \caption{ 
    \textbf{Comparison of different approaches for 
    detecting membership of AI Snake Oil.} 
    \textbf{Bold} indicates statistically significant results ($p < 0.05$).
    Our approach results in lower p-values compared 
    to other approaches (lower is better).
    }
    \begin{center}
        \begin{sc}
            \begin{tabular}{lc}
            \toprule
            Method & P-value ($\downarrow$) \\
            \midrule
            LLM DI \cite{maini2024llmdatasetinferencedid} (1) & 0.44 \\
            LLM DI \cite{maini2024llmdatasetinferencedid} (2) & 0.58  \\
            \midrule
            \datasetnamespace (w/o paired tests)    & 0.07    \\
            \datasetnamespace                       & \textbf{2.4e-3} \\
            \bottomrule
            \end{tabular}
        \end{sc}
        \label{table:blog_DI_results_appendix}
    \end{center}
\end{table}

\paragraph{Results.}

Table \ref{table:emnlp_DI_results_appendix} 
demonstrates \datasetname's effectiveness 
in detecting dataset membership for the blog articles.
LLM DI is unable to detect membership under the two 
different choices of validation set:
(1) with the private rephrases of the same $44$ blog posts as the validation set, and
(2) with the version of the \emph{held out} set of $12$ blog posts that is watermarking using the public key.
In Table \ref{table::mia_metrics} we evaluate
state-of-the-art MIAs and finding that they  
perform no better than random chance (AUROC $\approx$ 0.5).
Our findings corroborate with recent studies 
\cite{duan2024membership, maini2024llmdatasetinferencedid, das2024blindbaselinesbeatmembership}
that highlight the failure of sequence level MIAs on LLMs.

\section{Prompt Templates for Rephrasing} \label{appendix:prompt_template}

In this section, we outline the prompts used with 
LLaMA-3 70B \cite{llama3modelcard} to generate 
watermarked versions of the documents used in our 
experiments. 

\begin{tcolorbox}[colback=blue!8, colframe=gray!75, rounded corners, sharp corners=northeast, sharp corners=southwest,title=Prompt Template for Rephrasing Benchmarks,label=prompt:llama_prompt_benchmarks]
\texttt{Rephrase the question given below. Ensure you keep all details present in the original, without omitting anything or adding any extra information not present in the original question.\\ \\ \textbf{Question: }What is the main energy source for deep ocean currents that move large volumes of water around the planet?\\ \\Your response should end with "Rephrased Question: [rephrased question]"}
\end{tcolorbox}

\begin{tcolorbox}[colback=blue!8, colframe=gray!80, rounded corners, sharp corners=northeast, sharp corners=southwest,title=Prompt Template for Rephrasing Abstracts,label=prompt:llama_prompt_emnlp]
\texttt{Rephrase the abstract of a ML research paper given below following these strict guidelines:\\\\PRESERVE:\\- All technical details and findings\\- Original tone of the abstract\\\\AVOID:\\- Adding interpretive language not present in the original abstract\\- Removing any details\\- Changing meaning or emphasis\\\\Abstract: \{original\_abstract\}\\\\Your response should end with "Rephrased Abstract: \{rephrased\_abstract\}"}
\end{tcolorbox}

\begin{tcolorbox}[colback=blue!8, colframe=gray!80, rounded corners, sharp corners=northeast, sharp corners=southwest,title=Prompt Template for Rephrasing Blogs,label=prompt:llama_prompt_blog]
\texttt{Rephrase the below paragraph from an AI newsletter while maintaining coherent flow between paragraphs. Here are your instructions:\\\\1. I will provide the previous paragraph (marked as CONTEXT) and the current paragraph to rephrase (marked as TARGET).\\2.Your task is to:\\- Rephrase the TARGET paragraph so it flows naturally from the previous paragraph (CONTEXT)\\- Keep the same tone and emphasis as the original paragraph\\-Preserve the technical details present in the original paragraph\\- Do not add any extra information not present in the original paragraph\\- Avoid making sentences wordier or adding interpretive language\\\\3. Format your response as: REPHRASED PARAGRAPH: [your rephrased version]\\\\Context: \{context\}\\Paragraph: \{paragraph\}}
\end{tcolorbox}

\section{Watermarked Examples} \label{appendix:rephrased_examples}

\subsection{Watermarked Test Sets}  \label{appendix:rephrased_examples_benchmarks}

\subsubsection{TriviaQA}

\begin{tcolorbox}[colback=blue!8, colframe=gray!75, rounded corners, sharp corners=northeast, sharp corners=southwest]
\texttt{\textbf{Original Question:} Which enduring cartoon character was created by Bob Clampett for the 1938 cartoon Porky's Hare Hunt?\\ \\ \textbf{Rephrased Question:} Which long-lasting cartoon character was originally created by Bob Clampett for the 1938 cartoon titled 'Porky's Hare Hunt'?}
\end{tcolorbox}

\begin{tcolorbox}[colback=blue!8, colframe=gray!75, rounded corners, sharp corners=northeast, sharp corners=southwest]
\texttt{\textbf{Original Question:} Which US state lends its name to a baked pudding, made with ice cream, sponge and meringue?\\ \\ \textbf{Rephrased Question:} Which US state is the namesake of a baked pudding that consists of sponge, meringue, and ice cream?}
\end{tcolorbox}

\subsubsection{ARC Challenge}

\begin{tcolorbox}[colback=blue!8, colframe=gray!75, rounded corners, sharp corners=northeast, sharp corners=southwest]
\texttt{\textbf{Original Question:} Company X makes 100 custom buses each year. Company Y makes 10,000 of one type of bus each year. Which of the following is the most likely reason a customer would buy a bus from company X instead of company Y?\\ \\ \textbf{Rephrased Question:} What is the most probable reason a customer would choose to purchase a bus from Company X, which produces 100 custom buses annually, over Company Y, which manufactures 10,000 buses of a single type each year? }
\end{tcolorbox}

\begin{tcolorbox}[colback=blue!8, colframe=gray!75, rounded corners, sharp corners=northeast, sharp corners=southwest]
\texttt{\textbf{Original Question:} Sugars are necessary for human cell function. Which of the following are human cells not capable of doing?\\ \\ \textbf{Rephrased Question:} Given that sugars are necessary for human cell function, what is it that human cells are unable to do?}
\end{tcolorbox}

\subsubsection{MMLU}

\begin{tcolorbox}[colback=blue!8, colframe=gray!75, rounded corners, sharp corners=northeast, sharp corners=southwest]
\texttt{\textbf{Original Question:} Noradrenaline is the neurotransmitter between which of the two structures below?\\ \\ \textbf{Rephrased Question:} Between which two structures listed below does noradrenaline act as the neurotransmitter? }
\end{tcolorbox}

\begin{tcolorbox}[colback=blue!8, colframe=gray!75, rounded corners, sharp corners=northeast, sharp corners=southwest]
\texttt{\textbf{Original Question:} On which surfaces of the teeth is dental plaque most likely to accumulate in the mouth of a patient with poor oral hygiene?\\ \\ \textbf{Rephrased Question:} In a patient with poor oral hygiene, on which surfaces of the teeth is dental plaque accumulation most probable in the mouth? }
\end{tcolorbox}

\subsubsection{GSM8K}

\begin{tcolorbox}[colback=blue!8, colframe=gray!75, rounded corners, sharp corners=northeast, sharp corners=southwest]

\texttt{\textbf{Original Question:} Darrell and Allen's ages are in the ratio of 7:11. If their total age now is 162, calculate Allen's age 10 years from now.\\ \\ \textbf{Rephrased Question:} If the current ages of Darrell and Allen are in a 7:11 ratio and their combined age at present is 162, what will be Allen's age a decade from now. }
\end{tcolorbox}

\begin{tcolorbox}[colback=blue!8, colframe=gray!75, rounded corners, sharp corners=northeast, sharp corners=southwest]
\texttt{\textbf{Original Question:} Indras has 6 letters in her name. Her sister's name has 4 more letters than half of the letters in Indras' name. How many letters are in Indras and her sister's names?\\ \\ \textbf{Rephrased Question:} If Indras' name contains 6 letters, and her sister's name has a number of letters that is 4 more than half the number of letters in Indras' name, then how many letters are there in Indras' and her sister's names combined? }
\end{tcolorbox}

\subsection{Watermarked Paper Abstracts} \label{appendix:rephrased_examples_abstracts}

\begin{tcolorbox}[
colback=blue!8, 
colframe=gray!50, 
rounded corners, 
sharp corners=northeast, 
sharp corners=southwest,
title= \textcolor{blue}{Tree of Problems: Improving structured problem solving with compositionality \cite{zebaze2024treeproblemsimprovingstructured}} ]
\texttt{\textbf{Original Abstract:}\\\\ Large Language Models (LLMs) have demonstrated remarkable performance across multipletasks through in-context learning. For complex reasoning tasks that require step-by-step thinking, Chain-of-Thought (CoT) prompting has given impressive results, especially when combined with self-consistency. Nonetheless, some tasks remain particularly difficult for LLMs to solve. Tree of Thoughts (ToT) and Graph of Thoughts (GoT) emerged as alternatives, dividing the complex problem into paths of subproblems. In this paper, we propose Tree of Problems (ToP), a simpler version of ToT, which we hypothesise can work better for complex tasks that can be divided into identical subtasks. Our empirical results show that our approach outperforms ToT and GoT, and in addition per forms better than CoT on complex reasoning tasks. All code for this paper will be made available.\\ \\ \textbf{Watermarked Abstract:}\\\\ Large Language Models (LLMs) have achieved notable performance in multiple tasks using in-context learning. For complex reasoning tasks requiring step-by-step thinking, Chain-of-Thought (CoT) prompting, especially when combined with self-consistency, has yielded impressive results. However, certain tasks continue to pose significant challenges for LLMs. Tree of Thoughts (ToT) and Graph of Thoughts (GoT) have emerged as alternatives by breaking down complex problems into subproblem paths. This paper introduces Tree of Problems (ToP), a simplified ToT version that we hypothesize will be more effective for complex tasks comprising identical subtasks. Our empirical findings indicate that ToP outperforms both ToT and GoT, and also surpasses CoT on complex reasoning tasks. The code used in this paper will be made available. }
\end{tcolorbox}

\begin{tcolorbox}[
colback=blue!8, 
colframe=gray!50, 
rounded corners, 
sharp corners=northeast, 
sharp corners=southwest,
title= \textcolor{blue}{“A good pun is its own reword”: Can Large Language Models Understand Puns? \cite{xu-etal-2024-good}} ]
\texttt{\textbf{Original Abstract:}\\\\ Puns play a vital role in academic research due to their distinct structure and clear definition, which aid in the comprehensive analysis of linguistic humor. However, the understanding of puns in large language models (LLMs) has not been thoroughly examined, limiting their use in creative writing and humor creation. In this paper, we leverage three popular tasks, i.e., pun recognition, explanation and generation to systematically evaluate the capabilities of LLMs in pun understanding. In addition to adopting the automated evaluation metrics from prior research, we introduce new evaluation methods and metrics that are better suited to the in-context learning paradigm of LLMs. These new metrics offer a more rigorous assessment of an LLM’s ability to understand puns and align more closely with human cognition than previous metrics. Our findings reveal the “lazy pun generation” pattern and identify the primary challenges LLMs encounter in understanding puns.\\ \\ \textbf{Watermarked Abstract:}\\\\ Puns, due to their distinct structure and clear definition, play a vital role in academic research on linguistic humor analysis. Despite their importance, the understanding of puns in large language models (LLMs) remains understudied, which hinders their application in creative writing and humor generation. This paper employs three tasks - pun recognition, explanation, and generation - to systematically evaluate LLMs' capabilities in pun understanding. Building upon prior research's automated evaluation metrics, we propose new evaluation methods and metrics tailored to the in-context learning paradigm of LLMs. These novel metrics provide a more rigorous assessment of an LLM's pun understanding ability and better align with human cognition. Our results expose the "lazy pun generation" pattern and pinpoint the primary challenges LLMs face in understanding puns.}
\end{tcolorbox}

\begin{tcolorbox}[
colback=blue!8, 
colframe=gray!50, 
rounded corners, 
sharp corners=northeast, 
sharp corners=southwest,
title= \textcolor{blue}{{M}ini{C}heck: Efficient Fact-Checking of {LLM}s on Grounding Documents \cite{tang-etal-2024-minicheck} } ]
\texttt{\textbf{Original Abstract:}\\\\ Recognizing if LLM output can be grounded in evidence is central to many tasks in NLP: retrieval-augmented generation, summarization, document-grounded dialogue, and more. Current approaches to this kind of fact-checking are based on verifying each piece of a model generation against potential evidence using an LLM. However, this process can be very computationally expensive, requiring many calls to a model to check a single response. In this work, we show how to build small fact-checking models that have GPT-4-level performance but for 400x lower cost. We do this by constructing synthetic training data with GPT-4, which involves creating realistic yet challenging instances of factual errors via a structured generation procedure. Training on this data teaches models to check each fact in the claim and recognize synthesis of information across sentences. For evaluation, we unify datasets from recent work on fact-checking and grounding LLM generations into a new benchmark, LLM-AggreFact. Our best system MiniCheck-FT5 (770M parameters) outperforms all systems of comparable size and reaches GPT-4 accuracy. We release LLM-AggreFact, code for data synthesis, and models.\\ \\ \textbf{Watermarked Abstract:}\\\\ Verifying the evidential grounding of LLM outputs is crucial for various NLP tasks, including retrieval-augmented generation, summarization, and document-grounded dialogue. Existing fact-checking approaches typically involve verifying each component of model generation against potential evidence using an LLM, which can be computationally expensive and require multiple model calls for a single response. This work demonstrates the construction of compact fact-checking models that achieve GPT-4-level performance at a 400x lower cost. We accomplish this by generating synthetic training data using GPT-4 through a structured procedure that creates realistic yet challenging instances of factual errors. Models trained on this data learn to verify facts within claims and recognize information synthesis across sentences. We establish a unified benchmark, LLM-AggreFact, by consolidating datasets from recent fact-checking and LLM grounding research. Our top-performing system, MiniCheck-FT5 (770M parameters), outperforms comparable-sized systems and matches GPT-4's accuracy. We make LLM-AggreFact, the data synthesis code, and the models publicly available.}
\end{tcolorbox}

\end{document}